# Automating global landslide detection with heterogeneous ensemble deep-learning classification

Alexandra Jarna Ganerød [1,2*†], Gabriele Franch [3†], Erin Lindsay [4], Martina Calovi [1]

[1] NTNU, Department of Geography, Trondheim, Norway; alexandra.jarna@ntnu.no, martina.calovi@ntnu.no
[2] NGU, Geological Survey of Norway, Trondheim, Norway; alexandra.jarna@ngu.no
[3] FBK, Fondazione Bruno Kessler, Trento, Italy; franch@fbk.eu
[4] NTNU, Department of Civil and Environmental Engineering, Trondheim, Norway; erin.lindsay@ntnu.no

* Correspondence: alexandra.jarna@ntnu.no; Tel: +47 4512 6688
[†] Author 1 and Author 2 contributed equally to this work

**Highlights:**

- **Building a heterogeneous ensemble of models substantially improves the prediction performance.**
- **Ensemble learning methods reduced the uncertainty of the single models' results.**
- **The best result is achieved with using S1 & S2 only data (F1 = 0,66)**
- **Biggest improvement Setting 2 only S2 bands ensemble model increase from 0.52 to 0.61.**
- **Possibility of creating a monitoring system based only on Setting 2 (dNDVI & land cover classification) – notice of possible change in the area.**

## ABSTRACT

With changing climatic conditions, we are already seeing an increase in extreme weather events and their secondary consequences, including landslides. Landslides threaten infrastructure, including roads, railways, buildings, and human life. Hazard-based spatial planning and early warning systems are cost-effective strategies to reduce the risk to society from landslides. However, these both rely on data from previous landslide events, which is often scarce. Many deep learning (DL) models have recently been applied for landside mapping using medium- to high-resolution satellite images as input. However, they often suffer from sensitivity problems, overfitting, and low mapping accuracy. This study addresses some of these limitations by using a diverse global landslide dataset, using different segmentation models, such as Unet, Linknet, PSP-Net, PAN, and DeepLab and based on their performances, building an ensemble model. The ensemble model achieved the highest F1-score (0.69) when combining both Sentinel-1 and Sentinel-2 bands, with the highest average improvement of 6.87 % when the ensemble size was 20. On the other hand, Sentinel-2 bands only performed very well, with an F1 score of 0.61 when the ensemble size is 20 with an improvement of 14.59 % when the ensemble size is 20. This result shows considerable potential in building a robust and reliable monitoring system based on changes in vegetation index dNDVI only.

## Introduction

Landslides occur daily around the world and can cause significant damage to infrastructure and property. Human life and health can also be at stake. In recent years, their occurrence and associated costs have increased, given the increase in extreme precipitation events and unregulated urban expansion in landslide-prone areas (Froude and Petley, 2018; Gariano and Guzzetti, 2016; Hanssen-Bauer et al., 2009). However, landslides are frequently underrepresented in global natural catastrophe databases due to a lack of reporting or because insurance claims due to landslide damages may be categorised under the primary triggering event, such as earthquakes or extreme weather events such as hurricanes or floods (Kirschbaum et al., 2010).

The scarcity of data on previous landslide events is a major limiting factor in developing mitigating strategies for future risks posed by landslides (Guzzetti et al., 2012). Two cost-effective methods for mitigating landslide risk to society include hazard-based spatial planning of new developments and early warning systems (Piciullo et al., 2018; Shano et al., 2020). However, these both rely on data from previous landslide events. Recent technological developments offer great potential to improve landslide detection through automatic landslide monitoring systems using satellite imagery and deep learning-based image classification algorithms (Mondini et al., 2021; Tehrani et al., 2021). Since 2017, there has been a significant increase in research into this topic (Mondini et al., 2021).

Two types of satellite images are mainly used for automatic landslide detection. These are optical images, from which vegetation indices (commonly, the normalised difference vegetation index, NDVI) can be derived, and synthetic aperture radar (SAR) images. Change detection methods are often used, either from single pre- and post-event images or composites produced from multitemporal image stacks, that help reduce noise from seasonal changes in vegetation or snow, clouds, or speckle (Lindsay et al., 2022). Landslides are most clearly visible in optical change images; however, SAR imagery has the advantage of being cloud penetrating and, therefore, can detect landslides more rapidly in case of scarcity of cloud-free images. With optical images, it is common to see landslides based on a loss of vegetation, which produces a negative NDVI value (Lin et al., 2005). For the changes in ground surface texture and the associated scattering behaviour that is detected by using cross- or co-polarised backscatter intensity images, we are using SAR images (Lindsay et al., 2022). A detailed explanation of the physical mechanisms that control landslide expression in SAR backscatter change images is available (Lindsay et al., 2022). Geographical Information Systems (GIS) have helped with data preparation, but manual mapping remains a time-consuming and subjective practice. Therefore, there is considerable potential in using automated machine- and deep-learning techniques to increase data collection and make landslide mapping practice more efficient (Bai et al., 2022; Ganerød et al., 2023; Ghorbanzadeh et al., 2019; Nava et al., 2022; Prakash et al., 2021).

Computer-based methods are rapidly emerging as powerful, efficient, and viable tools able to work with high-dimensional datasets (Chen and Lin, 2014; Dargan et al., 2020; Zhu et al., 2017). These methods can thus reduce costly and time-consuming manual labour both in the field and during post-digital interpretation. Applying computer-based methods to the landslide susceptibility maps can play a crucial role in determining the most vulnerable and prone areas for landslides (Shano et al., 2020; Tehrani et al., 2022; Wei et al., 2022; Youssef and Pourghasemi, 2021). The well-mapped landslides are the primary key to further developing a reliable monitoring system. And since landslide hazard mapping relies on knowledge gained from previous events (Band et al., 2020; Mondini et al., 2021; Prakash et al., 2021), the availability of high-quality data can help in developing strategies that aim to reduce landslide hazards and consequently, the risk to society.

The literature shows that qualitative and quantitative methods can be applied to detect landslides (Saadatkhah et al., 2014). Data-driven machine-learning techniques are widely used nowadays, but they still present some weaknesses that affect the predictive performances of single models (Lv et al., 2022). The scarcity of training data gives typical weaknesses. Most DL models require a large amount of diverse training data. This issue can cause the possibility of missing the best-fit function during the training process (Blum, A., Wang, 2011; Thomas G. Dietterich, 1997; Wang et al., 2011). However, considering ensemble learning in landslide applications, the majority of the studies are conducted for landslide susceptibility mapping (Bai et al., 2022; Hong et al., 2020; Lin et al., 2022; Liu et al., 2021; Lv et al., 2022; Merghadi et al., 2020; Saha et al., 2021; Setargie et al., 2023; Tong et al., 2023; Wei et al., 2022) rather than landslide detection or prediction.

With this study, we compare different segmentation models using PyTorch, in combination with two different learning rates (0,001 and 0,01) and five loss functions to test how the prediction and statistics change using different settings and Fully Convolutional Neural Networks (FCNN). Out of 30 well-mapped study areas, (Lindsay et al., 2023) presented as potential case studies, 21 representing the descending orbit were selected for our study. Limitations are connected to the landslides on east-facing slopes that are more likely to be detectable in descending SAR data due to increased sampling density on slopes facing towards the sensor and geometric distortions. In contrast, the total number of case studies with descending obit was higher. Furthermore, we evaluate whether the resulting predictions vary according to the different settings: (1) only Sentinel-1, (2) only Sentinel-2, (3) combinations of Sentinel-1 and Sentinel-2 and lastly, (4) all the bands by calculating precision, recall, F1-score on the test dataset. Based on all combinations, the different ensemble learning settings are assessed to calculate the performance of each ensemble model based on the 10, 20 and 40 best-performing single models on validation data. The goal of this study is threefold: first, to find the setting with the best performance and least data amount. Second, to create an ensemble model, compare its performance and evaluate its average improvement to the single models. Third, assess the resulting performances and discuss their potential use.

## Case studies

This study uses the global landslide datasets created by Lindsay et al. (2023b), where landslides have been manually mapped mainly through dNDVI images derived from Sentinel-2. This dataset included 30 case studies worldwide, characterised by varying terrain types, orientation and size, ground cover, climate zones, geological materials, and failure mechanisms. These 30 sites have been mainly identified thanks to the reports titled 'The Landslide Blog' (https://blogs.agu.org/landslideblog/), news reports, or journal articles found through an extensive online search. The locations and dates of these mapped landslide events are shown in Figure 1. In contrast, the specific properties of each event, together with the local environmental conditions, are shown in Table 1 (Lindsay et al., 2023).

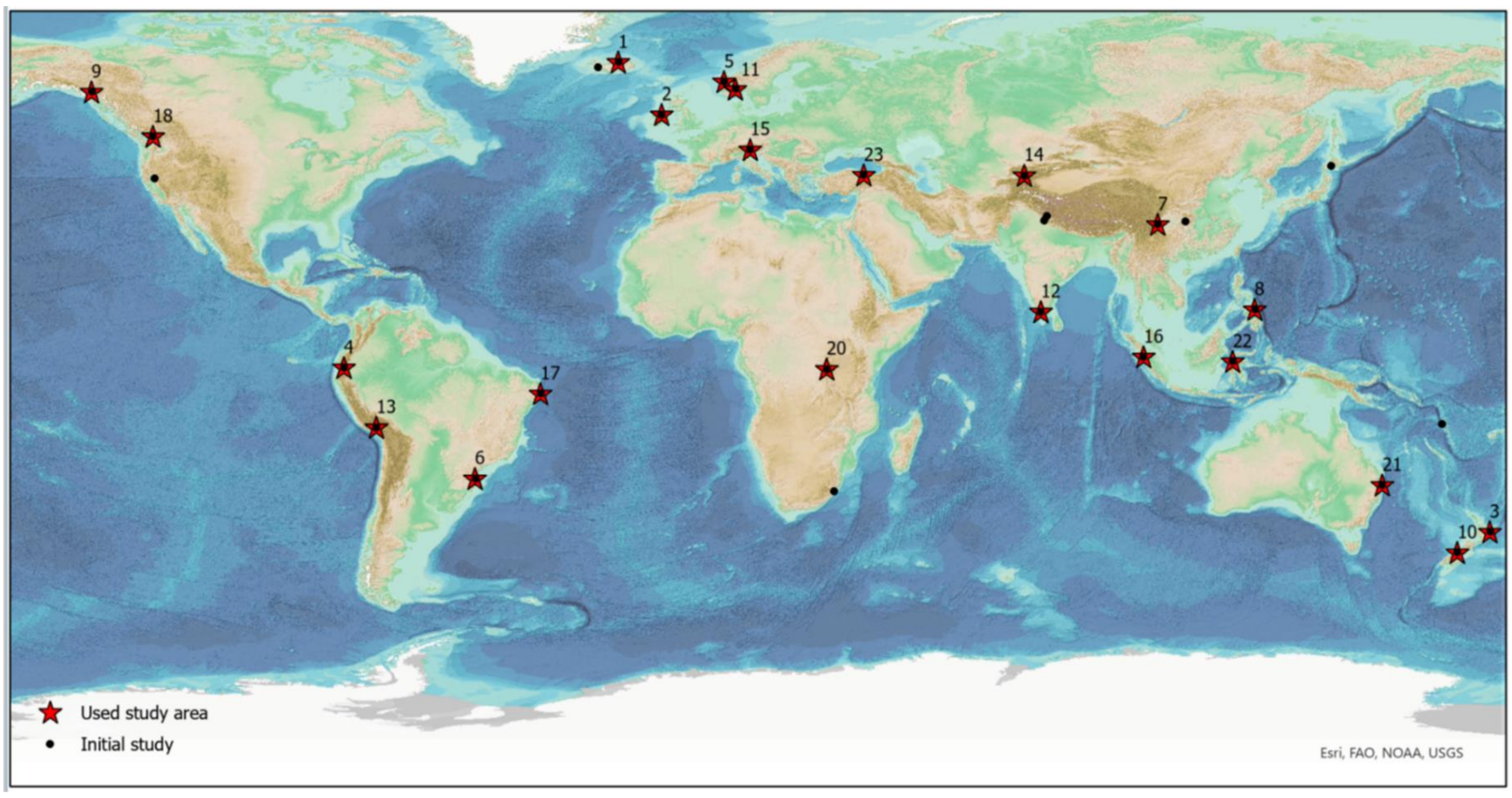

| | | |
|---|---|---|
| 1. Iceland | 2021-10-06 | (-17.5558, 65.8674) |
| 2. Ireland | 2021-07-04 | (-7.85187, 54.15745) |
| 3. New Zealand | 2022-03-27 | (177.4694, -38.9681) |
| 4. Ecuador | 2021-02-12 | (-78.93655, -2.2895) |
| 5. Norway | 2019-07-30 | (6.11699, 61.51734) |
| 8. Brazil | 2020-04-06 | (-49.60777, -27.0774) |
| 9. China | 2021-04-05 | (103.15055, 29.66474) |
| 10. Phillipines | 2022-04-10 | (124.85286, 10.67658) |
| 12. USA | 2019-12-02 | (-135.41194, 59.22035) |
| 14. New Zealand | 2022-03-27 | (170.17485, -43.57436) |
| 15. Iceland | 2022-02-02 | (-22.109811, 64.787941) |
| 19. India | 2020-08-06 | (77.01188, 10.166) |
| 20. Peru | 2020-06-16 | (-71.685998, -15.660252) |
| 21. Kyrgzstan | 2017-04-30 | (73.290135, 40.587722) |
| 22. Italy | 2018-11-02 | (11.972334, 46.34617) |
| 23. Indonesia | 2022-02-25 | (99.99442, 0.11412) |
| 24. Brazil | 2022-05-26 | (-34.958299, -8.120723) |
| 25. Canada | 2021-11-14 | (-121.63403, 49.307154) |
| 26. USA | 2021-09-24 | (-121.38802, 39.86029) |
| 28. Australia | 2022-02-28 | (153.39578, -28.51734) |
| 29. Indonesia | 2018-09-28 | (119.91418, -0.93951) |

Figure 1. Lindsay et al., 2023 case study locations and event dates (Lindsay et al., 2023). The case studies where landslides were not visible or did not have descending data have been removed—base map: Landslide Hazard Map (GFDRR, 2023).

**Table 1.** Lindsay et al., 2023: 21 used case studies. **Type** refers to the different types of landslides, specifically: 1. Rockfall, 14. Clay/silt compound slide, 22. Debris flow, 23. Mudflow, 25. Debris avalanche, 26. Earthflow, 27. T refers to the trigger: R – rainfall, E- earthquake, S – snowmelt, U - unknown. Size presents length and width in kilometres, Rainfall in millimetres per year and occurrence date.

| Location | Type | T | Size L x W [km] | Rainfall [mm/yr] | Date |
|---|---|---|---|---|---|
| **1. Iceland*** | 22 | R | 0.8 x 0.1 | 672 | 06.10.2021 |
| **2. Ireland** | 27 | R | 0.58 x 0.7 | 1358 | 04.07.2021 |
| **3. N. Zealand** | 13 | R | 0.13 x 0.05 | 1508 | 27.03.2022 |
| **4. Ecuador** | 26 | R | 1.5 x 1.5 | 918 | 12.02.2021 |
| **5. Norway*** | 22 25 | R | 0.11 x 0.03 | 2285 | 30.07.2019 |
| **8. Brazil** | 22 | R | 1.6 x 0.02 | 1547 | 06.04.2020 |
| **9. China** | 13-22 | R | 1.34 x 0.92 | 1297 | 05.04.2021 |
| **10. Philippines** | 23 | R | 2.1 x 0.7 | 2915 | 10.04.2022 |
| **12. USA** | 25 | R | 1.7 x 0.18 | 1282 | 02.12.2019 |
| **14. N. Zealand** | 18 | R | 1.8 x 0.28 | 4222 | 27.03.2022 |
| **15. Iceland** | 18-25 | R | 2.4 x 1.7 | 829 | 02.02.2022 |
| **19. India** | 22 | R | 1.2 x 0.12 | 2848 | 06.08.2020 |
| **20. Peru** | 26 | U | 0.6 x 1 | 506 | 16.06.2020 |
| **21. Kyrgyzstan** | 14-26 | RS | 5 x 0.6 | 394 | 30.04.2017 |
| **22. Italy** | 22 | R | 0.35 x 0.07 | 886 | 02.11.2018 |
| **23. Indonesia** | 22 | E | 6 x 0.3 | 2775 | 25.02.2022 |
| **24. Brazil** | 13 | R | 0.06 x 0.03 | 1678 | 26.05.2022 |
| **25. Canada** | 18 | R | 0.85 x 0.32 | 1712 | 14.11.2021 |
| **26. USA** | 1 | U | 0.09 x 0.06 | 1560 | 24.09.2021 |
| **28. Australia** | 22 | R | 0.8 x 0.04 | 2031 | 28.02.2022 |
| **29. Indonesia** | 16 | E | 2.1 x 1.1 | 1534 | 28.09.2018 |

In this study, we included only the case studies with Sentinel-1 – 1 descending orbit images to have the highest amount of use cases with the same parameters (21 cases). Only the visible landslides in both Sentinel-1 and Sentinel-2 bands (Figure 2) have been included for the final evaluation of each single model. We classified them as follows: The Norway case (#5) was reserved as a test and used to evaluate the area. In contrast, the remaining 20 cases have been randomly split into 16 locations for model training and 4 locations for model validation. This split was performed only once and has been used for all the experiments (fixed random seed). Figure 2 shows the landslides mapped in this area's Norwegian (#5) evaluation set.

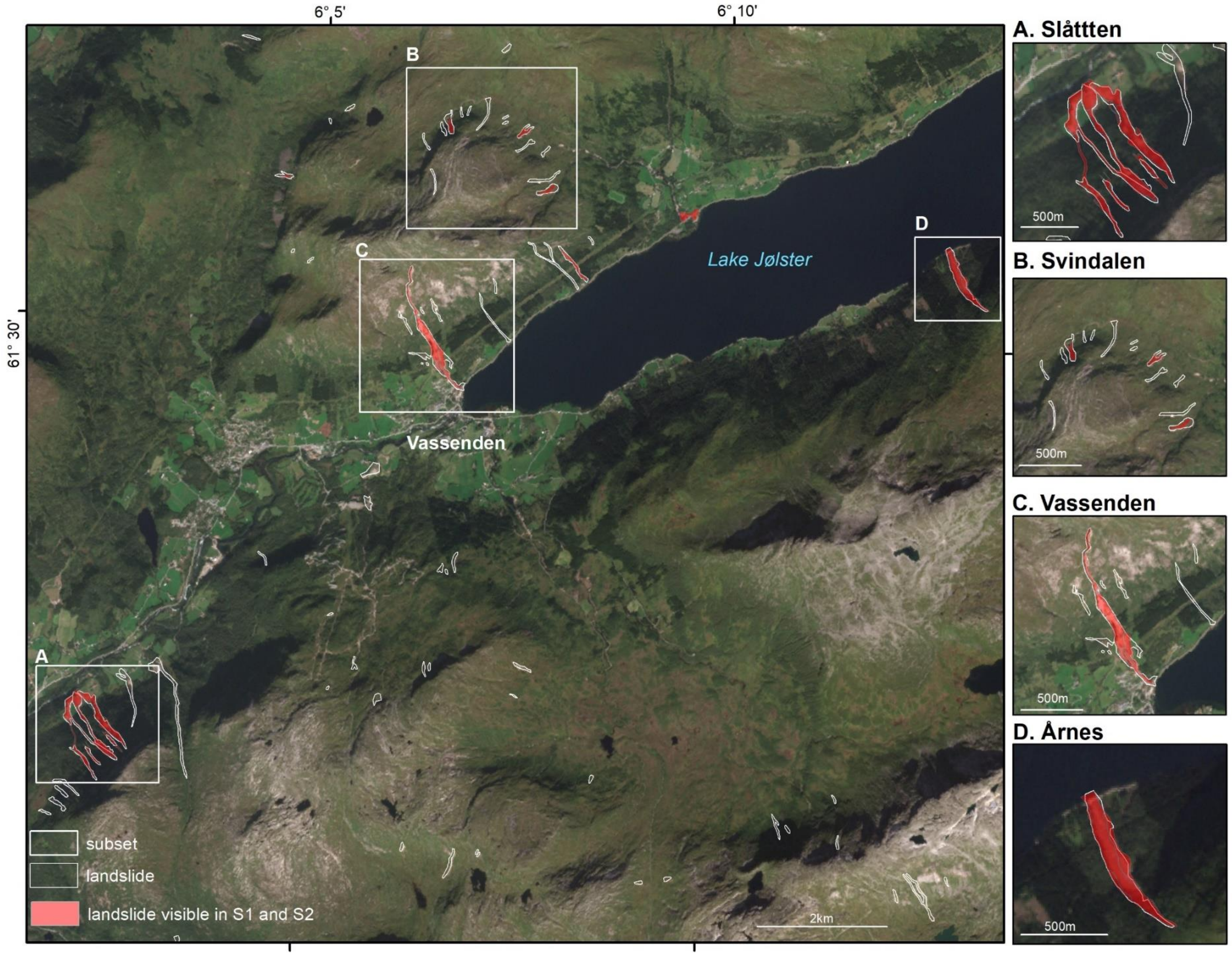

Figure 2: Norwegian case study area #5. Norway is the test area, showing four subsets with ground truth landslide outlines in white. However, only red-defined regions have been used for test evaluation because they are visible in both Sentinel-1 and Sentinel-2 layers.

Looking at the size of the case study areas, eight case studies show less than 0.1 km$^2$, and five were between 0.1 and 1 km$^2$. The remaining eight events are between 1 and 5 km$^2$ (Table 1). The mean annual rainfall varied from 394 mm/yr in Kyrgyzstan (#21) to 4.222 mm/yr in New Zealand (#14). The land cover types, identified using the Copernicus Global Land Cover classification map, included 7 cases of forest and 6 of herbaceous, cropland, or shrubs. For most of our case studies, the main trigger events that initiated the landslide are rainfall (16 events), earthquakes (2 events), a combination of rainfall and snowmelt (1 event only), and in 2 cases, the triggers are unknown (see Table 1).

## Methods – data and computing

### 3.1 Input data – bands

This section describes the input data and the pre-processing, visualisation, and interpretation methods, summarised by Lindsay et al. (2023). The Sentinel-1 SAR change images and time series were prepared for each case study using Google Earth Engine (GEE) (Gorelick et al., 2017). The date of occurrence is crucial because it defines the correct ranges for the pre-and post-event image collections. The coordinates of the approximate event location have been used as filter conditions to produce pre- and post-event image stacks of both Sentinel-1 GRD (Ground Range Detected) products and Sentinel-2 (Level 2A) images. Periods of one, two, or 12 months before or after the event date were used, ensuring comparison between temporally homogenous periods, depending on the local image acquisition frequency and seasonality. The date ranges and coordinate locations for each case study are shown in Figure 1. Pre- and post-event image composites were produced from the image stacks for a 4 km2 area about the defined point for all the cases. The Sentinel-1 images, retrieved from GEE, are available pre-processed (calibrated and ortho-corrected) at 10 m resolution. A terrain correction (Vollrath et al., 2020) was applied to each image in the stack using either a volumetric or a surface model, depending on the land cover type. The terrain correction projects the side-looking images onto the terrain while masking out pixels with shadow, and layover corrects over- and under-brightening. The 30 m resolution Shuttle Radar Topography Mission (SRTM) Digital Elevation Model (DEM), available within GEE, was used for this. However, local 10 m DEMs were utilised for the Icelandic case (link: ÍslandsDEM v1.0 10m - awesome-gee-community-catalog, accessed: 11 December 2022) and for Norwegian patients (https://hoydedata.no/LaserInnsyn2/, accessed: 11 December 2022).

Furthermore, Sentinel-1 composites were created by taking the median of the terrain-corrected image collections and changing images (post minus pre-event image composite). For Sentinel-2, an NDVI band was added to each image in the pre-and post-event stacks; then, a greenest-pixel composite was created (maximum NDVI) using the quality-mosaic tool. We produced an NDVI change image (dNDVI) as a final product by subtracting the pre- from the post-event composite. The dNDVI images are the only Sentinel-2-band included in the dataset (Figure 3).

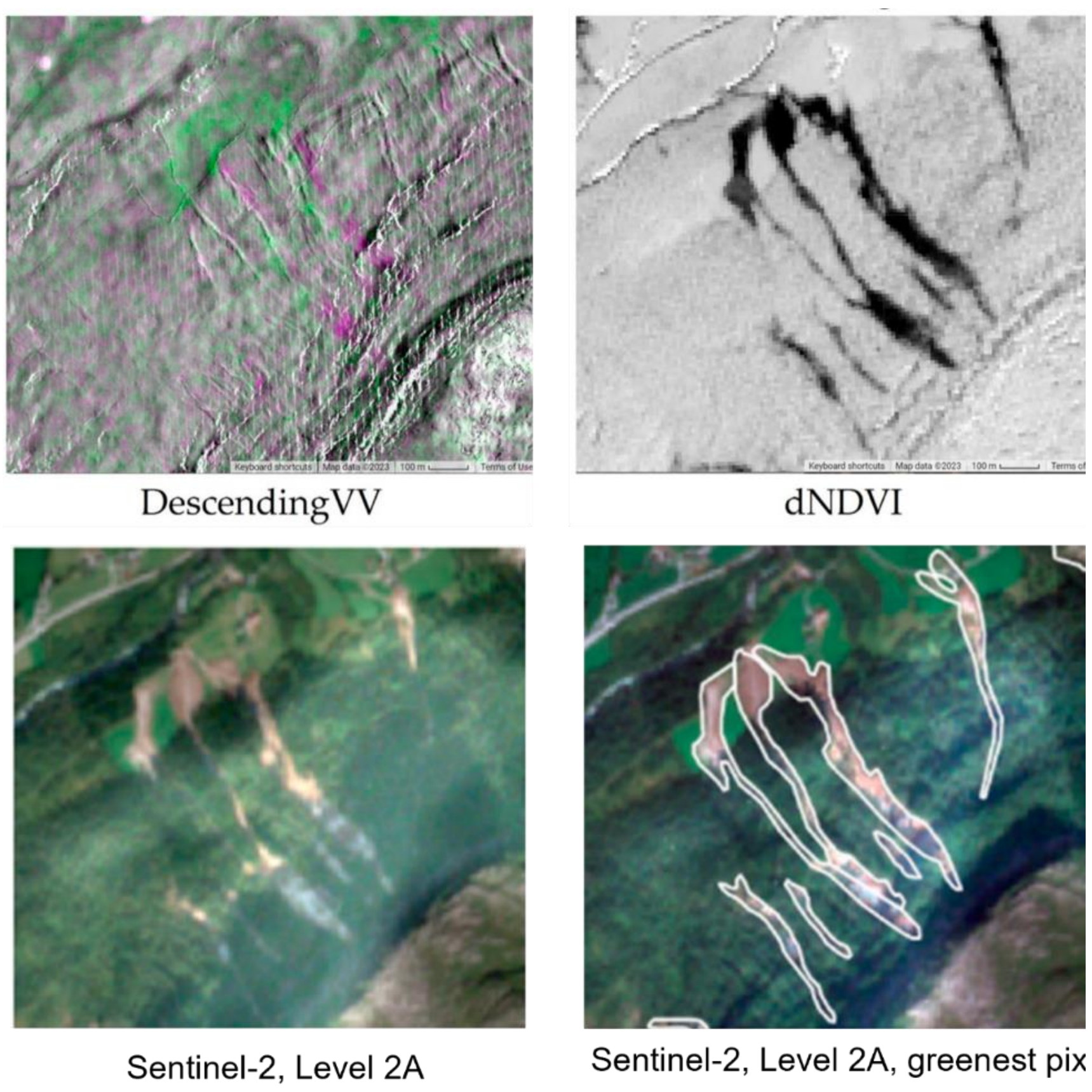

Figure 3: Example of the landslide showing case study #5 (test case study as a subset) in (a) multi-temporal VV-polarized SAR backscatter descending intensity change images (green indicates a backscatter intensity increase; purple indicates a decrease); (b) Differenced Normalised Difference Vegetation Index (dNDVI); (c) Sentinel- Level 2A, with atmospheric correction applied to the Level-1C TOA image. (d) cloud-filtered, greenest-pixel composite produced from Sentinel-2 Level2A images (white outlines were mapped from the Sentinel-2 dNDVI image).

Table 2 Bands description. Acronyms: pre- pre-event, post – post-event image. V – vertical polarisation, H – horizontal polarisation; Digital Elevation Model (DEM)

| Band | | Resolution [m] | Origin | Description |
|---|---|---|---|---|
| **0** | **preVV** | 10 | Sentinel-1 + DEM | Pre-event VV polarised backscatter intensity. VV is good for showing the difference in surface roughness. |
| **1** | **postVV** | 10 | Sentinel-1 + DEM | Post-event VV polarised backscatter intensity. |
| **2** | **dVV** | 10 | Sentinel-1 + DEM | PostVV minus PreVV |
| **3** | **preVH** | 10 | Sentinel-1 + DEM | Pre-event VH polarised backscatter intensity. VH is more sensitive to changes in biomass. |
| **4** | **postVH** | 10 | Sentinel-1 + DEM | Post-event VH polarised backscatter intensity. VH is more sensitive to changes in biomass. |
| **5** | **dVH** | 10 | Sentinel-1 + DEM | PostVH minus PreVH |
| **6** | **Layover** | 10 | Sentinel-1 + DEM | Mask of layover distortion areas, calculated from (Vollrath et al., 2020) terrain correction. It should match the NoData areas in the VV and VH bands. |
| **7** | **Shadow** | 10 | Sentinel-1 + DEM | Mask of shadow distortion areas, calculated from the Vollrath et al. (2020) terrain correction. It should match the NoData areas in the VV and VH bands. |
| **8** | **liaDeg** | 10 | Sentinel-1 + DEM | The local incidence angle (LIA) is in degrees. The angle describes the orientation of the ground surface relative to the sensor's line of sight (Lindsay et al., 2023). |
| **9** | **Elevation** | 30 | DEM | The SRTM global 30 m resolution DEM was used for case studies below 60 deg latitude. Above 60 deg, other local DEMs were used – Icelandic 10 m and Norway 10 m resolution. |
| **10** | **Slope** | 30 | DEM | Derived from elevation |
| **11** | **KG_climate** | | | Köppen-Geiger climate map with a spatial resolution of 5 arc minutes for 1986-2010. http://koeppen-geiger.vu-wien.ac.at/ |
| **12** | **Popatpv_tree height** | 30 | LiDar | A 30-m spatial resolution global forest canopy height map - through the integration of the Global Ecosystem Dynamics Investigation (GEDI) LIDAR forest structure measurements and Landsat analysis-ready data time series (Landsat ARD) (Potapov et al., 2021). |
| **13** | **CART_LC_classified** | 10 | Sentinel-1 and Sentinel-2 | Land cover classification. This was prepared for each case study individually, based on Sentinel-1 and -2 images from pre-event, using a CART machine-learning algorithm in Google Earth Engine. |
| **14** | **dNDVI** | 10 | Sentinel-2 | Change in Normalised Difference Vegetation Index (NDVI) |

The Local Incidence Angle (LIA) (liaDeg in Table 2) was included to investigate the factors that influence the expression of landslides in the change images. The LIA is particularly important when there has been a height change due to erosion of soil or rock material or the removal of trees (O'Grady et al., 2013).

Here, we used the manually mapped landslide polygons for each case study, exporting the shapefiles along with the Geotiff raster images for each case study, and finally extracting and plotting the pixel values from within the polygons.

For land cover classification (CART_LC_classified), locally trained land cover maps were produced. A machine-learning-based land cover classification was performed for each case study using the ee. smile.CART algorithm (Classification and Regression Tree) in GEE (Breiman et al., 2017). These were trained using a Sentinel-2 image with minimal cloud cover from before the event and the DEM and pre-event Sentinel-1 composite images (VV and VH polarisation) (Lindsay et al., 2023). The classifier was trained by manually selecting points within the following classes: urban, artificial; forest, trees, scrub, herbaceous; pasture, grass; sparsely vegetated; water body, sea; wetland; bare rocks; and glacier and snow (Lindsay et al., 2023).

Finally, for each case study, the following were exported: 1) landslide polygons (reclassified as a binary geotiff), (2) geotiff raster images, including the pre-processed Sentinel-1 bands from only descending mode, were used for training. The pixel values within each of the landslide polygons were extracted to produce a dataset consisting of approximately 300,000 pixels. As a result, from the initial 30 case studies, only 21 have been included in this study.

### 3.2 Proposed settings

All the presented bands are in logical content and calculated with the use of DEM, Sentinel-1, or Sentinel-2 data. Here we decided to apply our combination of segmentation models, loss functions, and learning rate (Section 3.3) as follows (Table 3): Setting 1 is proposed to use nine bands created with the use of only Sentinel-1 data, while Setting 2 presents two bands of multi-temporal dataset containing only Sentinel-2 data. Setting 3 combines both Sentinel-1 and Sentinel-2 bands and Setting 4 includes all 15 multi-temporal bands.

Table 3. Proposed and tested settings 1-4 in this study.

| Setting | | Bands |
|---|---|---|
| **1** | only Sentinel-1 | 0,1,2,3,4,5,6,7,8 |
| **2** | only Sentinel-2 | 13,14 |
| **3** | Sentinel-1 and Sentinel-2 | 0,1,2,3,4,5,6,7,8,13,14 |
| **4** | all bands | 0,1,2,3….14 |

### 3.3 Segmentation models and loss functions

Image segmentation is a crucial task in computer vision and image processing. It has various applications, including scene understanding, medical image analysis, robotic perception, video surveillance, augmented reality, and image compression (Blaschke, 2010; Blaschke et al., 2014). The widespread success of deep learning (DL) has led to the development of image segmentation approaches that utilise DL models (Al-Obeidat et al., 2016; Minaee et al., 2022) in many fields of application. In our study, we build upon these foundations. Specifically, we relied on the models provided by the open-source library segmentation models PyTorch (SMP - https://smp.readthedocs.io/), which implements nine different segmentation models (many of which have shown state-of-the-art results when presented). The aim was to evaluate them both in single model configuration and ensemble settings to analyse the performances for the geospatial dimension in landslide mapping with high imbalance in the dataset (Table 4).

Loss functions guided deep learning model training to optimise model parameters (weights). The loss function compared the ground truth and predicted output values, and the model weights were optimised to minimise the loss values between the predicted and target outputs. In our work, we trained all nine models on five different loss functions among all losses provided by the SMP library: BCE, Dice, Focal, Jacckard, and Lovasz Loss (Table 5). The loss functions were chosen as these are known to be a good fit for semantic segmentation problems (Jadon, 2020).

The last configuration we explore for model training is the learning rate parameter, where we test both 0.01 and 0.001 as possible values. To recap, we trained nine deep learning model architectures (Table 4) in combination with five loss functions (Table 5): BCELoss, DiceLoss, FocalLoss, JaccardLoss, LovaszLoss, and two learning rates of 0.01 and 0.001, which in total resulted in 9x5x2 = 90 combinations (training sessions) for each Settings 1-4 (Table 3). The best-performing models on the validation set among these 90 were further used to form the multi-model ensembles.

Table 4. Segmentation models. All the models are Fully Convolutional Neural Networks (FCNN) with the same supported metadata: classified tiles and task: pixel classification.

| Segmentation Models | Stands for | Categories (Minaee et al., 2022) | Specialty | Source |
|---|---|---|---|---|
| Unet | U shape | Encoder-decoder (3) | works with fewer training images | (Ronneberger et al., 2015) |
| Unet++ | U shape ++ | Nested encoder-decoder (3) | more complex decoder | (Zhou et al., 2018) |
| MA-Net | Multi-scale Attention Net | Multiscale and Pyramid Network-Based Model (4) | Two blocks: position-wise and multi-scale fusion attention block | (Fan et al., 2020) |
| Linknet | LinkNet | Encoder-decoder (3) | uses sum for fusing decoder blocks | (Chaurasia and Culurciello, 2018) |
| FPN | Feature Pyramid Network | Multiscale and Pyramid Network-Based Model (4) | without the need to compute image pyramids<br>- suitable for small objects | (Li *et al.*, **2019**) |
| PSP-Net | Pyramid Scene Parsing Network | Multiscale and Pyramid Network-Based Model (4) | not suitable for small objects-based context aggregation. | (Zhao et al., 2017) |
| PAN | Pyramid Attention Network | The Regional CNN (R-CNN) (5) | data-to-text generation | (Jiang et al., 2020) |
| DeepLabV3 | DeepLab version 3 | Dilated Convolutional Model (6) | uses dilated convolutions | (Chen et al., 2018a) |
| DeepLabV3+ | DeepLab version 3+ | Dilated Convolutional Model (6) | adds a simple yet effective decoder module | (Chen et al., 2018b) |

Table 5. Loss functions used in this study all support binary, multiclass, and multilabel. Here, we describe their specifics.

| Loss function | Specifics |
|---|---|
| BCELoss | creates a criterion that measures the Binary Cross Entropy between the target and the input probabilities. |
| DiceLoss | a standard metric for pixel segmentation that can also be modified to act as a loss function |
| FocalLoss | applies a modulating term to the cross-entropy loss in order to focus learning on complex misclassified examples. It is a dynamically scaled cross-entropy loss, where the scaling factor decays to zero as confidence in the correct class increases. |
| JaccardLoss | similar to the Dice metric and is calculated as the ratio between the overlap of the positive instances between two sets and their mutual combined values |
| LovaszLoss | It is designed to optimise the Intersection over Union score for semantic segmentation, particularly for multi-class instances. |

### 3.4 Training, validation and model selection

Data augmentation (Ghasemi et al., 2022) is a technique for artificially increasing the training set by creating modified copies of a dataset from the same training data. Given the relatively small size of our training set (16 images of different sizes), we employ multiple augmentation options to increase the data diversity as much as possible. Among many augmentation techniques, in our experiment, we choose to employ both “smart” random cropping with a resolution of 256x256 pixels and random 90-degree rotation. The “smart” random cropping selects a random area of the image to contain at least one positive pixel (a pixel marked as a landslide); this is done to reduce the data imbalance and guarantee that every input image always contains both labels.

We employ an early stopping technique for all training sessions to avoid overfitting. Specifically, we train the model for a fixed number of steps (1000 iterations with batch size 4, forming 1 training epoch). Then, we evaluate the classification performance on the validation set (4 images) by computing the F1 score metric (see section 3.5 for the F1 score definition). When the F1 score no longer improves for 50 epochs, we stop training and keep the models’ weights with the best F1 score on the validation set as the final model for testing.

### 3.5 Evaluation and ensemble model

Model evaluation was performed quantitatively by computing the confusion matrix between the label and the model’s classification on the Norwegian test set (Figure 2). Table 6 shows the equations for the performance metrics (precision, recall, and F1-score, all ranging from 0 to 1) computed from confusion, starting from true positive (TP), false positive (FP), false negative (FN), and true negative (TN) values. We used the F1 score as the main performance metric given its suitability in highlighting classification performance differences on datasets with unbalanced classes (with 0 indicating a model with no skill and 1 indicating a perfect skill).

Table 6. Equations for performance evaluation metrics from confusion matrix values.

| Metric | Formula |
|---|---|
| **Precision** | $\frac{TP}{TP + FP}$ |
| **Recall** | $\frac{TP}{TP + FN}$ |
| **F1-score** | $\frac{2TP}{2TP + FP + FN}$ |

All metrics are summarised in the performance diagram (Roebber, 2009) and presented in Figure 5, which shows precision, recall, F1 score and frequency bias in a single plot, highlighting both the overall performance and if the model or ensemble is overpredicting or underpredicting respect to the ground truth.

As introduced in section 3.3, we trained nine segmentation architectures with five loss functions and two learning rates on four different settings, creating a total of 360 models (9x5x2x4) with other statistical proprieties. In statistics and machine learning, ensemble methods are commonly used to combine multiple model outputs to obtain better predictive performance than could be obtained from any of the constituent learning algorithms on their own (Band et al., 2020; Yang et al., 2022). We built and tested ensembles of different sizes for all four settings by averaging the best-performing models' classification results on the validation set (Table 7). Figure 4 graphically summarises the ensemble prediction process:

- The K best models (2 <= K <= 90) with the highest score on the validation set are collected to form the ensemble.
- Input data is processed individually by each model in the ensemble, generating K output classification maps.
- The pixel average of the K classification maps is computed as the final classification output.

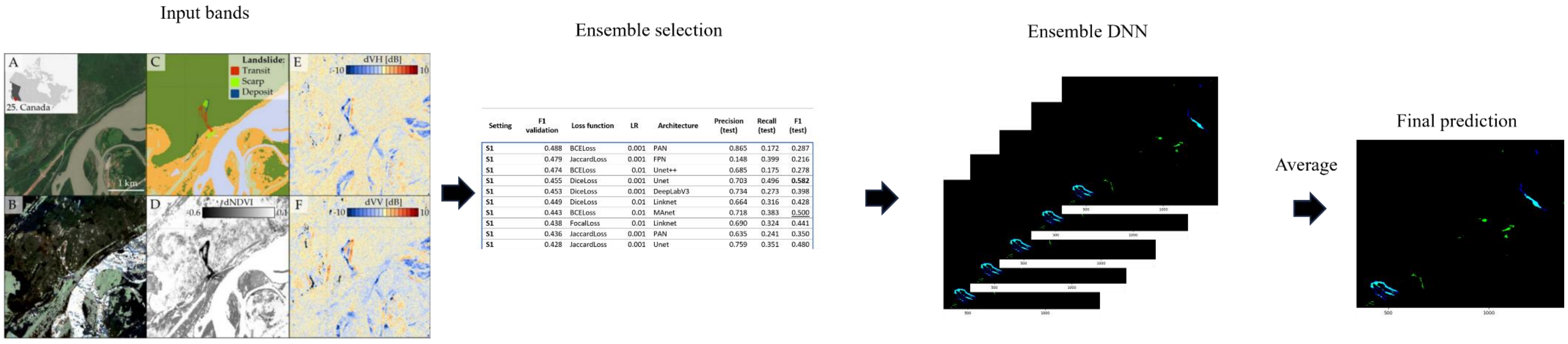

Figure 4: Workflow of compiling the final prediction using ensemble techniques. The best-performing models' output is averaged to obtain the final prediction.

## 4. Results

In this section, we analyse the outcomes of all the four settings, both on a single model perspective and with ensembles of different sizes focused on the automated detection of landslides using an ensemble global deep-learning approach.

### 4.1 Single model performance

Table 7 summarises the characteristics (model architecture, loss function, learning rate) of the ten best-performing combinations of each Setting 1-4. On average, lower scores on the validation translate into lower performances on the test set, indicating a positive correlation between training/validation and test performance. Thus, the models were correctly trained to generalise without overfitting. The worst performing setting is Sentinel 1 (S1), where the best validation F1 scores range between 0.428 and 0.488, with a corresponding range in test performance between 0.216 and 0.582, with 5 out of 10 models with a test score below 0.4. Sentinel 2 (S2) shows substantially improved validation scores, from 0.637 to 0.694 and consistently more stable test performances, from 0.42 to 0.56. The best performances overall are given by the settings using both Sentinel 1 and Sentinel 2 bands (S1+S2) with a very consistent high validation scores (0.741 to 0.751) and test performance (from 0.581 to 0.695). Counterintuitively, the additions of other bands on top of S1 and S2 in the last setting (all) result in a slight degradation in performance in validation (0.706 to 0.727) and mixed results in test (0.591 to 0.687), hinting that the additional bands may be of limited or detrimental utility for landslide identification.

Interestingly, in the best-performing setting, S1+S2, 7 out of 10 models use the Unet++ architecture. All the other locations instead show a varied mix of model, loss and learning rate combination in the top 10, highlighting that most architectures were able to deliver good performances given the proper configuration.

Table 7: Ten best-performing segmentation models for each setting based on validation F1 score and their corresponding performance metrics on the test set. In **bold,** the best performance on the test set underlined the worst performance.

| Setting | F1 validation | Loss function | Learning Rate | Architecture | Precision (test) | Recall (test) | F1 (test) |
|---|---|---|---|---|---|---|---|
| **S1** | 0.488 | BCELoss | 0.001 | PAN | 0.865 | 0.172 | 0.287 |
| **S1** | 0.479 | JaccardLoss | 0.001 | FPN | 0.148 | 0.399 | 0.216 |
| **S1** | 0.474 | BCELoss | 0.01 | Unet++ | 0.685 | 0.175 | 0.278 |
| **S1** | 0.455 | DiceLoss | 0.001 | Unet | 0.703 | 0.496 | **0.582** |
| **S1** | 0.453 | DiceLoss | 0.001 | DeepLabV3 | 0.734 | 0.273 | 0.398 |
| **S1** | 0.449 | DiceLoss | 0.01 | Linknet | 0.664 | 0.316 | 0.428 |
| **S1** | 0.443 | BCELoss | 0.01 | MAnet | 0.718 | 0.383 | 0.500 |
| **S1** | 0.438 | FocalLoss | 0.01 | Linknet | 0.690 | 0.324 | 0.441 |
| **S1** | 0.436 | JaccardLoss | 0.001 | PAN | 0.635 | 0.241 | 0.350 |
| **S1** | 0.428 | JaccardLoss | 0.001 | Unet | 0.759 | 0.351 | 0.480 |
| **S2** | 0.694 | FocalLoss | 0.01 | Linknet | 0.516 | 0.534 | 0.525 |
| **S2** | 0.691 | BCELoss | 0.01 | Unet++ | 0.322 | 0.673 | 0.435 |
| **S2** | 0.687 | JaccardLoss | 0.001 | PAN | 0.553 | 0.510 | 0.531 |
| **S2** | 0.682 | BCELoss | 0.01 | Linknet | 0.355 | 0.514 | 0.420 |
| **S2** | 0.668 | DiceLoss | 0.01 | Unet++ | 0.297 | 0.747 | 0.425 |
| **S2** | 0.645 | JaccardLoss | 0.01 | DeepLabV3 | 0.516 | 0.490 | 0.503 |
| **S2** | 0.643 | LovaszLoss | 0.001 | DeepLabV3+ | 0.495 | 0.646 | **0.560** |
| **S2** | 0.641 | JaccardLoss | 0.001 | Unet | 0.342 | 0.679 | 0.455 |
| **S2** | 0.640 | FocalLoss | 0.01 | MAnet | 0.375 | 0.730 | 0.496 |
| **S2** | 0.637 | BCELoss | 0.01 | Unet | 0.352 | 0.690 | 0.467 |
| **S1+S2** | 0.751 | FocalLoss | 0.01 | Unet++ | 0.708 | 0.589 | 0.643 |
| **S1+S2** | 0.750 | FocalLoss | 0.001 | Unet++ | 0.647 | 0.598 | 0.621 |
| **S1+S2** | 0.748 | JaccardLoss | 0.01 | Unet++ | 0.791 | 0.620 | **0.695** |
| **S1+S2** | 0.747 | DiceLoss | 0.001 | Linknet | 0.806 | 0.589 | 0.681 |
| **S1+S2** | 0.746 | JaccardLoss | 0.001 | FPN | 0.709 | 0.492 | 0.581 |
| **S1+S2** | 0.746 | DiceLoss | 0.001 | Unet++ | 0.636 | 0.616 | 0.626 |
| **S1+S2** | 0.745 | JaccardLoss | 0.001 | Unet++ | 0.678 | 0.624 | 0.650 |
| **S1+S2** | 0.744 | BCELoss | 0.01 | Unet++ | 0.716 | 0.546 | 0.619 |
| **S1+S2** | 0.742 | DiceLoss | 0.01 | Unet++ | 0.648 | 0.661 | 0.655 |
| **S1+S2** | 0.741 | BCELoss | 0.01 | Unet | 0.584 | 0.612 | 0.597 |
| **all** | 0.727 | DiceLoss | 0.001 | Unet | 0.710 | 0.599 | 0.650 |
| **all** | 0.725 | DiceLoss | 0.001 | Unet++ | 0.742 | 0.639 | **0.687** |
| **all** | 0.720 | JaccardLoss | 0.001 | PAN | 0.657 | 0.575 | 0.613 |
| **all** | 0.719 | JaccardLoss | 0.01 | FPN | 0.684 | 0.620 | 0.650 |
| **all** | 0.719 | JaccardLoss | 0.001 | Unet | 0.847 | 0.550 | 0.667 |
| **all** | 0.715 | JaccardLoss | 0.001 | Linknet | 0.812 | 0.556 | 0.660 |
| **all** | 0.713 | DiceLoss | 0.001 | FPN | 0.629 | 0.556 | 0.591 |
| **all** | 0.712 | DiceLoss | 0.01 | Unet | 0.788 | 0.568 | 0.660 |
| **all** | 0.708 | FocalLoss | 0.001 | Unet | 0.731 | 0.582 | 0.648 |
| **all** | 0.706 | JaccardLoss | 0.01 | Unet++ | 0.770 | 0.565 | 0.652 |

### 4.2 Ensemble Performance

Figure 5 and Table 8 summarise the performance improvement on the test set of building an ensemble of different sizes by averaging the predictions of the best-performing validation models. All four settings showed a sensible improvement in F1 skill when compared to the single best validation model. The skill improvement was more pronounced for the sets that performed lower in the single model configurations and shows a consistent pattern where most of the skill improvement was reached when considering ensemble sizes of up to 20 ensemble members. With the exclusion of the S1 setting, the use of ensemble sizes larger than 20 members determines a shift toward underestimation.

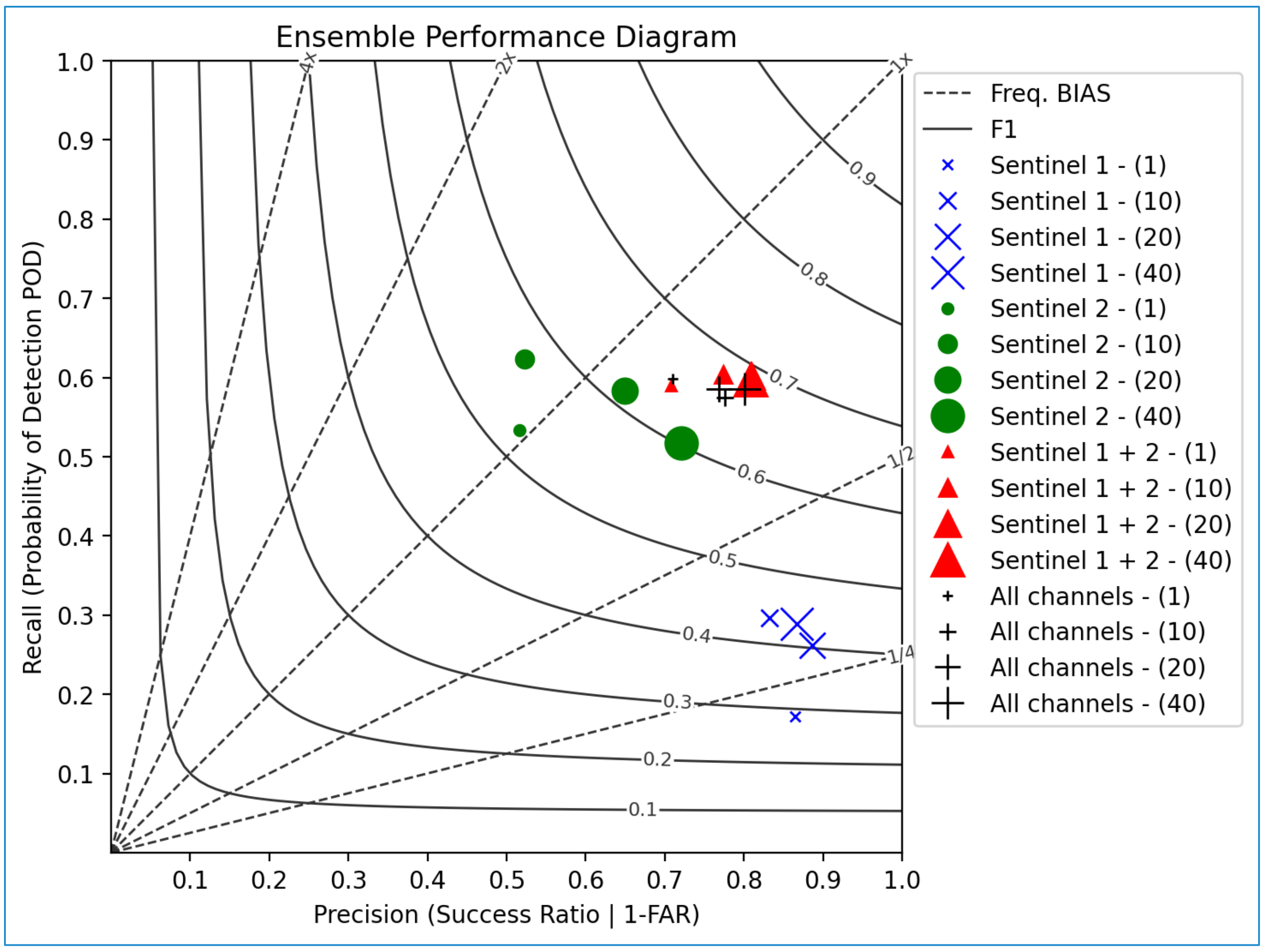

Figure 5: Performance diagram showing performance scores for all four settings on the test set. The model with the best validation score is compared with the ensemble average of the top 10, 20 and 40 models. Horizontal and vertical axes show precision and recall, respectively, while the curved lines represent the isoline of the F1 score. Dashed lines represent the frequency bias: if the marker is on the middle diagonal, the model outputs the same number of landslide pixels as the ground truth, if above the model is overestimating, if below, the model is underestimating. The closer the marker is to the top right corner, the better the classification performance (higher F1 score).

Table 8: Summarizing the ensemble performance diagram showing the F1 score for single model performance and performances of best 10 (F1 ens (10)), 20 (F1 ens (20)), and 40 (F1 ens (40)). The best-performing configuration is shown in bold, underlined with the second best, while in parenthesis, the % of improvement of the ensemble is compared to the single best model.

| | **Performance on test set (Improvement %)** | | | |
|---|---|---|---|---|
| **Setting** | **F1 single** | **F1 ens (10)** | **F1 ens (20)** | **F1 ens (40)** |
| **1) S1** | 0.29 | **0.44 (+34%)** | 0.40 (+29%) | 0.43 (+34%) |
| **2) S2** | 0.52 | 0.57 (+8%) | **0.61 (+15%)** | 0.60 (+13 %) |
| **3) S1+S2** | 0.64 | 0.68 (+5%) | **0.69 (+7%)** | 0.69 (+6 %) |
| **4) all** | 0.65 | 0.66 (+2%) | 0.66 (+2%) | **0.68 (+4 %)** |

Among the diverse configurations, our findings confirm that in ensemble settings, the most promising outcome was achieved by utilising both Sentinel-1 and Sentinel-2 bands (Figure 5 and Table 8). The statistics showed mostly differences in the overall performance using the different data settings when using only Sentinel-1 bands, where the overall F1 model performance based on the validation dataset shows the best score to be 0.44 for an ensemble model size of 10. On the other hand, Setting 2 with only two bands, achieved a high performance of 0.61 and an improvement of the ensemble model of 14.59% when the size is 20 models. For the combination of bands based on Sentinel-1 and Sentinel-2, the highest score achieved is equal to 0.69 for both ensemble sizes of 20 and 40. These observations are shown in Figure 6. The highest score was 0.68 for ensemble size 40 when using all the bands. This can also be observed when comparing the predictions to the reference datasets (Tables 8 and 9).

### 4.3 Performance Enhancement through Ensemble Models and Optimal Ensemble Size

Notably, the ensemble model employing only Sentinel 2 bands displayed the highest increase in performance using ensemble size 20. This means the viability of developing a monitoring system primarily based on Sentinel 2 bands, enabling the quick identification of potential changes in susceptible regions. Our analysis indicates that the most favourable results were attained when employing an ensemble size between 10 and 20 best-performing single models for Setting 2 and 4 (only S2 and all bands). However, we achieved the best performance for Settings 1 and 3 (only S1 and a combination of S1 and S2) when the ensemble size was between 5 and 10 single models. Beyond this threshold, a discernible shift toward underestimation became evident, suggesting a potential trade-off between ensemble size and accuracy. Figure 6 synthesises our findings, showing that after a certain size, adding more members does not improve, or is detrimental to, the overall skill due to the poor performance of the newly added models. As such, we suggest considering multi-model ensembles formed with up to 20 members.

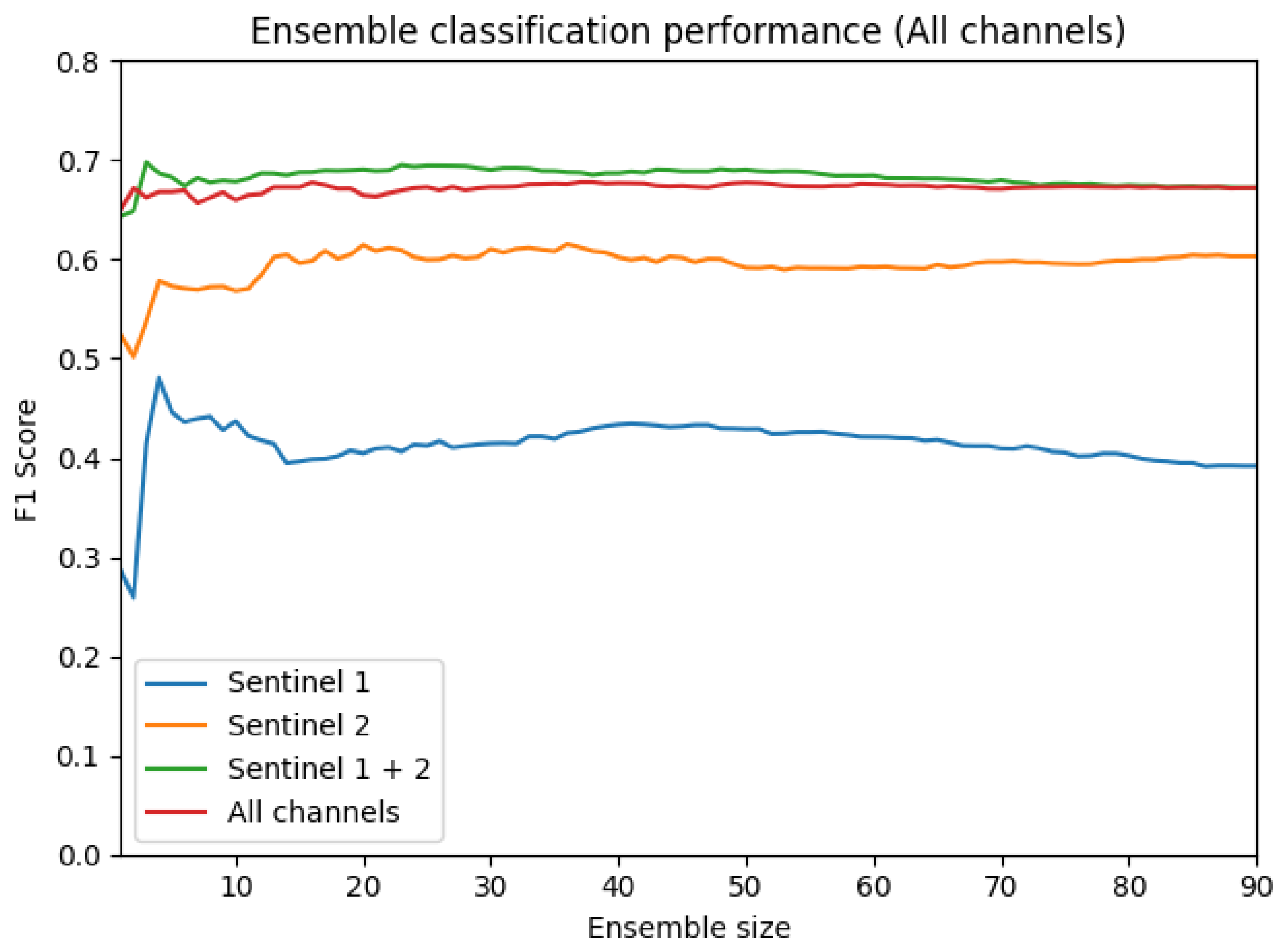

Figure 6: Plot showing the performance of all four settings when using an increasing number of models to build the ensemble one.

Lastly, we present the results as a set of four maps, presenting all four Settings. We created those maps based on the predictions generated through the use of the ensemble size 20 for all Settings. The respective performance information shown in Figure 7A with the colour cyan, where the prediction of a single model is missed. Still, the ensemble model of size 20 predicted the landslide correctly, and white colour represents the areas where both prediction and ensemble model predicted the landslide correctly. For Setting 1 (S1), we can see that only one of the large landslides was mapped correctly (represented in white on the map).

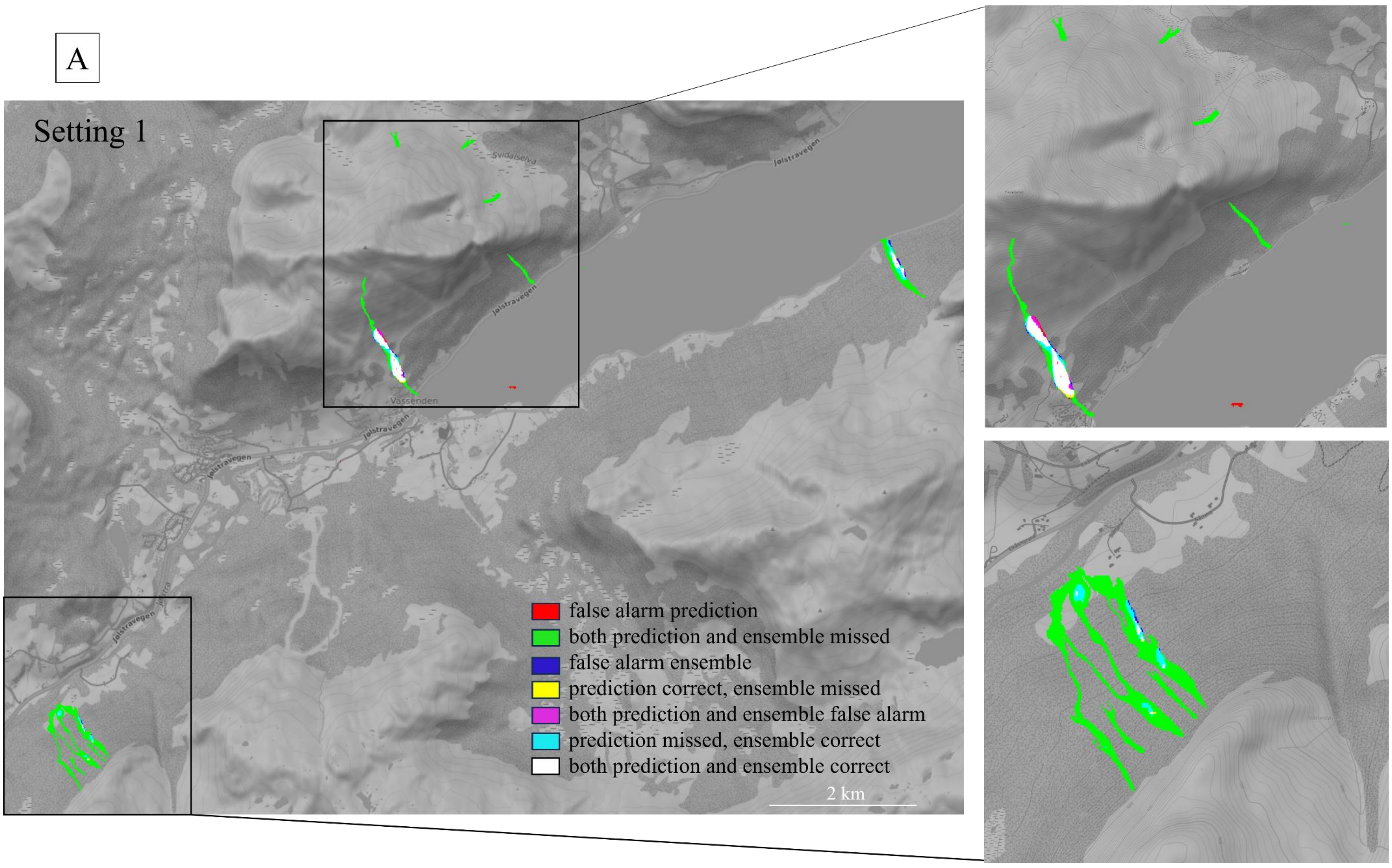

However, when applying Setting 2 (S2) (see Figure 7B), we can clearly see that most of the large landslides were detected by both the single and the ensemble model. The important and interesting fact in this case is that when we are looking at the test area (#5) shown in the Figure 2, many of the landslides represented in red and magenta correspond to the existing landslides. However, these were not used to evaluate the test scores because of the lack of visibility of these landslides within the Sentinel-1 bands. There are also false positives in the water, which do not correspond to landslides (red colour). Viewing the original dNDVI map, one can see that there can be areas of water that have a similar signature to landslides (negative NDVI), and the models here were not able to capture the difference. Map B confirmed that these landslides are only visible in Sentinel-2 bands, and in fact this setting was able to predict them. Also, the improvement of ensemble models – highlighted with in cyan, can be clearly seen in this figure both on the overview map and close-ups.

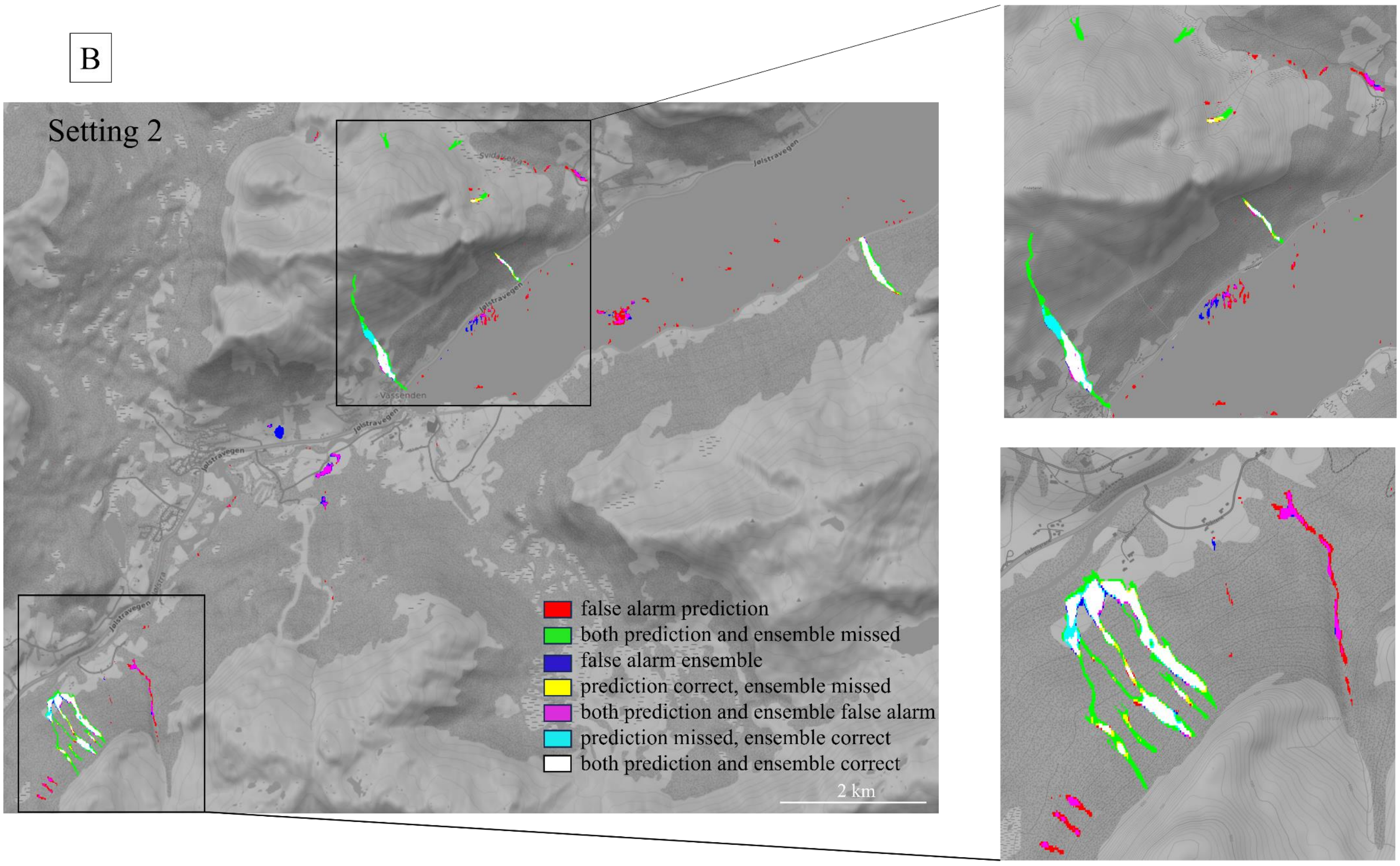

Setting 3 (S3) (Figure 7C) confirms that the best performance is achieved with the combination of the Sentinel-1 and Sentinel-2 bands, where both prediction of the single model and ensemble model predicted correctly many more the landslides.

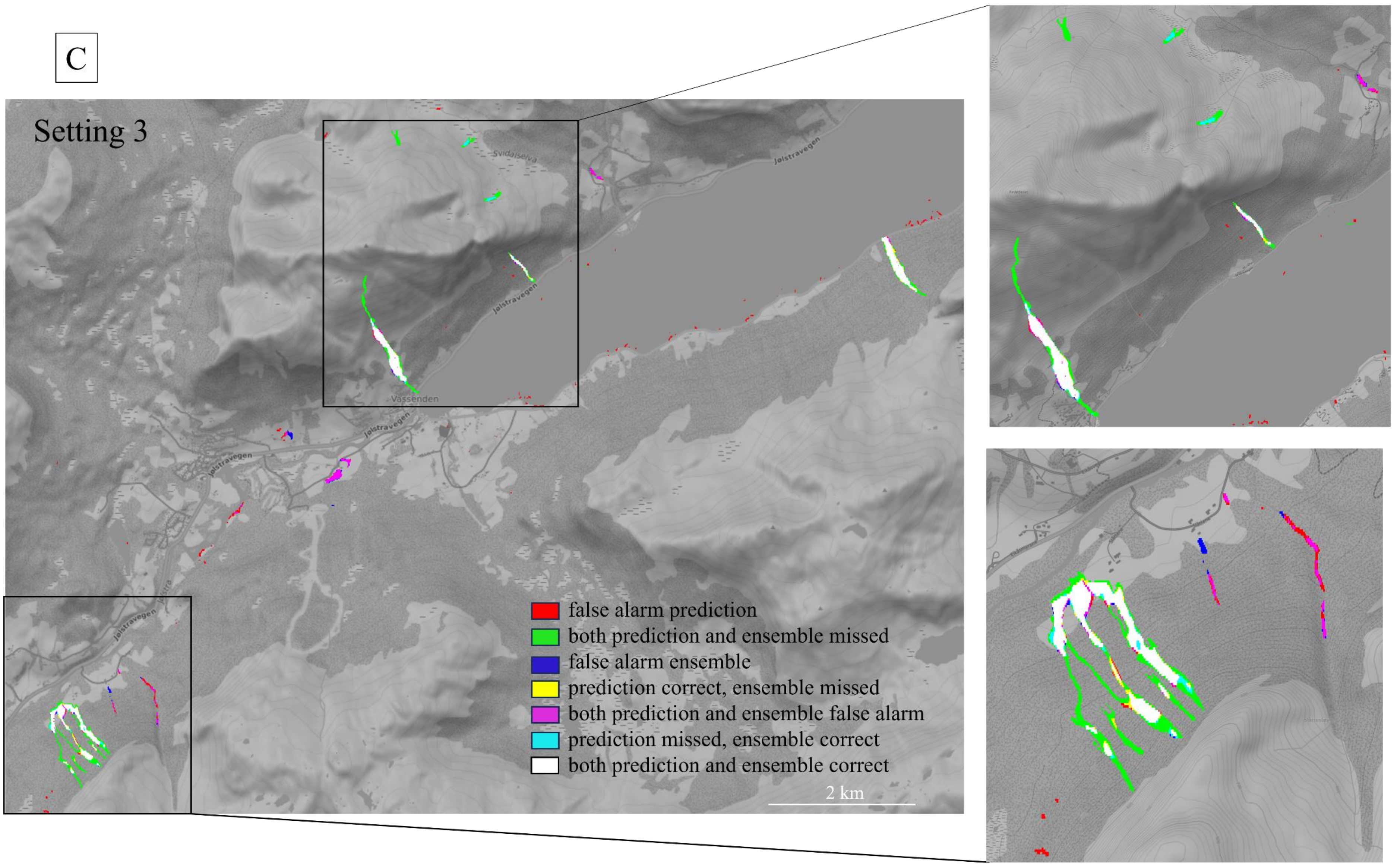

A similar phenomenon as in S2 can be seen in the map presenting Setting 4 (S4). The Sentinel-2 bands were present in training for S3 and S4 (Figure 7C, D). However, much less of red and pink was predicted compared to S2, where Sentinel-2 data stand-alone without any disturbance of other bands.

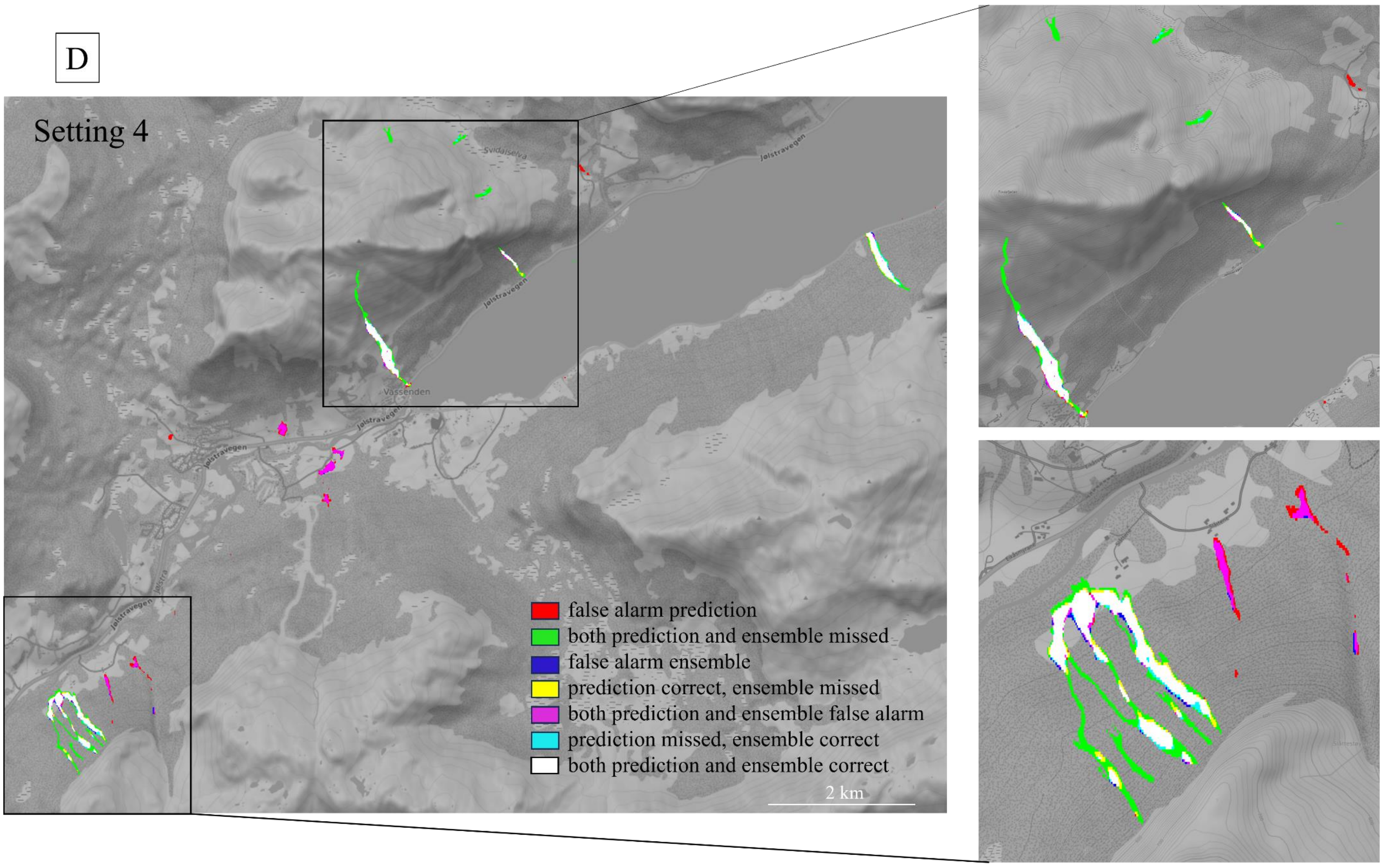

Figure 7: Test area showing the entire test area and two close ups to the landslide areas. For all Settings, the ensemble model is based on the size of 20 ensemble models. A) Setting 1 (Sentinel-1 only); B) Setting 2 (Sentinel-2 only); C) Setting 3 (combination of Sentinel-1 and Sentinel-2) and D) all the bands. (Colours meanings: red = false alarm prediction, green = both prediction and ensemble missed, blue = false alarm ensemble, yellow = prediction correct ensemble missed, magenta = both prediction and ensemble false alarm, cyan = prediction missed, ensemble correct, white = both prediction and ensemble correct)

## 5. Discussion

This study presents an automated approach for a globally trained deep-learning model for detecting landslides and testing the performance of four different settings based on Sentinel-1, Sentinel-2 and terrain data. The overarching objective of this study is to investigate the functionalities of varying segmentation models and evaluate which are the most suitable for automatic landslide detection. The results could potentially be used to develop a landslide monitoring system able to detect landslide and mitigate their hazards and risks to society. A lack of landslide data is a crucial limitation for the development of mitigation measures, including hazard mapping and early warning systems. The landslide spatial distribution and the local hazard conditions can be easily identified by using an automating detection system based on freely available images. Using a globally trained ensemble model approach helps overcome the lack of available training data. The one used in this study utilises a dataset that was originally designed to include an environmentally diverse range of case studies. The objective is to develop a model that could be used to predict landslides occurrences in a diverse range of environmental conditions. The case studies used for this project belong to a complete collection of different type of landslides occurred across the world over a period that goes from September 2018 to May 2022.

The performances of 10, 20 and 40 ensemble models were compared and presented. All the models' predictions performances were evaluated for recall, precision and F1 score, that set the choice of the ideal predictors (Ganerød et al., 2023; Ji et al., 2020; Liu et al., 2022).

The computational requirements for training the samples and running the model using deep learning techniques are not in the capacity of every computer. We tested the possibilities of a local machine with a highly powerful GPU (Nvidia RTX 4090 GPU (24GB VRAM)) running all the 360 single models and all the ensemble models. In total, four days (corresponding to 96 hours) were necessary to run the 360 models, averaging 16 minutes to train one single model. The training settings and the correct data loading enabled the GPU to its fully capabilities. Afterwards, the interference time to produce the prediction for the ensemble model the size 10, 20 and 40 on the test area was almost instantaneous. Therefore, a powerful GPU on the locally based computer proved to be working efficiently for this study. Furthermore, our study has unveiled novel insights into the effective detection and classification of landslides, significantly contributing to enhance the geohazard management research field. This discussion section presents our findings, the broader context of our research, and the potentials for further directions and work.

### 5.1 Combining SAR and optical data for a continuous monitoring system

The primary objective of our research was to advance the capabilities of landslide detection through the implementation of ensemble global deep-learning models. Our results indicate that the integration of diverse datasets, particularly the combination of Sentinel-1 and Sentinel-2 bands, achieved the highest values of F1 and reliability scores in landslide prediction. These results support the findings of a similar study that focuses on illegal logging detection (Luigi and Guzzetti, 2016). The highest performance, achieved by combining Sentinel-1 and Sentinel-2 data, was presented previously, where Sentinel-1 data improved overall landslide prediction. The locally trained deep learning models, such as the Unet, can accurately detect landslides from Sentinel-1 images. This is because these models can distinguish random speckle noise from clusters of pixels that indicate changes in the ground surface (Ganerød et al., 2023). This finding has significant implications into the rapid automatic detection field and can potentially be used as a part of a continuous monitoring system. When used in combination with Sentinel-2, the Sentinel-1 data helps to reduce false positives related to water bodies that occur when using Sentinel-2 data only. Using only Sentinel-2 data returns an overall higher detection rate. However, the key limitation in using it for a monitoring system is the necessary availability of cloud-free images, which can take several months in the worst cases. Another challenge that these models will face is related to the ability of distinguishing between signals of vegetation loss unrelated to landslides and those caused by landslides (Lindsay et al., 2023). This issue has persisted in landslide detection using optical data, as mentioned by Prakash et al. (2021). For a continuous monitoring system, we recommend to further develop a SAR-based detection model, which can be updated when cloud-free optical images become available.

### 5.2 Sentinel-2 for Landslide Inventory Production

One of the objectives of this study was to find the minimal amount of band with the highest performance. Therefore, we further evaluated combination of only Sentinel-2 bands in Setting-2 of the pre- from the post-event composite images. This approach can potentially be used for continuous landslide monitoring or as a tool for producing landslide inventories after a known rain event in the affected area (Lindsay et al., 2020). We believe that the dNDVI method is helpful in collecting more information about the size and location of landslides. However, it is not advised to use it in areas that covered by snow due to the lack of vegetation and changes in snow cover, which can lead to wrong predictions. There is a need for more investigation in differentiating landslides from changes in the snow. This suggests that inventories of landslides could be improved, which is important for defining thresholds and analysing the risks associated with them. (Ghorbanzadeh et al., 2019; Lindsay et al., 2020)

### 5.3 Optimal Ensemble Size

The testing of ensemble size's impact on performance revealed an intricate trade-off between the performance and the computational efficiency. While larger ensemble sizes demonstrated improved performance up to a certain point, a noticeable shift towards underestimation occurred as the ensemble size exceeded 20 models. These findings highlight the significance of carefully optimising ensemble size to strike a balance between precision and resource consumption.

### 5.4 Implications for Geohazard Management and Future Directions

Building on the current study, several future research can emerge. Exploring the integration of other remote sensing data sources and fine-tuning ensemble parameters could lead to even more accurate prediction models. Additionally, investigating the scalability and applicability of our findings to different geographic regions could further validate the robustness of the proposed approach. Based on our overall findings, we can recommend the implementation of a single setting and model architecture combination. Specifically, we suggest training the Unet++ with S1+S2 bands, given its consistent performance and robustness concerning parameter changes (loss function, learning rate).

## 6. Conclusions

Developments in computing, deep learning algorithms, and the increased availability of high-quality and free satellite data can potentially automate many mapping problems in Earth sciences. Choosing the most efficient machine learning algorithm is necessary to reduce inconsistencies in landslide detection. Usually, the main objective is to identify the optimal model based on its predictive capabilities. The primary aim of this study was to estimate the optimal model with maximum predictive ability, considering the limited availability of historical landslide data. We applied four different settings, all based on either Sentinel-1, Sentinel-2 and terrain model bands and their combinations, with nine segmentation models, two learning rates and five loss functions. In total, we generated 360 prediction models. The proposed framework is based on Python and uses PyTorch utilities. We used 21 case studies globally spread around the world. Our deep learning approach proved to delineate landslides with high performance in all the mentioned cases. However, the best-performing setting was based on a combination of Sentinel-1 and Sentinel-2 bands with the F1 score equal to 0.69. We achieved a promising performance of 0.6 when using Sentinel-2 bands only and applying the change in vegetation index (pre-and post-images) with the size of ensemble model 20. We observed that the Sentinel-2 only based ensemble model could delineate many more landslides and pick up those impossible to map using Sentinel-1 bands only. However, the main drawbacks of this approach were that in the test case results, it wrongly predicted landslides in the water, and it is necessary to wait for cloud-free images, which can limit the use for rapid detection. These findings help to inform the development of continuous monitoring systems and approaches for producing precise landslide inventories after widespread triggering events.

To summarise, our study underscores the potential of ensemble global deep-learning models in automating landslide detection. The integration of various data sources and the identification of context-specific applicability presents the potential for innovative geohazard management strategies. The outcomes also emphasise the significance of ensemble size, with an optimal range of 10 to 20 models providing the best compromise between calculated scores and computational efficiency for Settings 2 and 4 (only Sentinel-2 and all bands). However, we achieved the best performance for Settings 1 and 3 (only Sentinel-1 and a combination of Ssentinel-1 and Sentinel-2) when the ensemble size was between 5 and 10 single models. The insights are working towards developing robust early warning systems to minimise landslide hazards and their impact on society.

## Code availability section

The source code is available for download here: GitHub

Program language: Python

Software required: data preparation (ArcGIS Pro/QGIS)